\documentclass[10pt,twocolumn,letterpaper]{article}

\usepackage{multirow}
\usepackage[accsupp]{axessibility}
\usepackage{cvpr}  
\definecolor{cvprblue}{rgb}{0.21,0.49,0.74}
\usepackage[pagebackref,breaklinks,colorlinks,allcolors=cvprblue]{hyperref}
\usepackage{orcidlink}

\def\paperID{146} 
\def\confName{CVPR}
\def\confYear{2026}

\title{FSM-Net: An Efficient Frequency-Spatial Network for Real-World Deblurring}

\author{Vinh-Thuan Ly\orcidlink{0009-0005-7557-8231}\\
University of Science, VNU-HCM, Ho Chi Minh City, Vietnam\\
Vietnam National University, Ho Chi Minh City, Vietnam\\
{\tt\small 22280092@student.hcmus.edu.vn}
}

\begin{document}
\maketitle
\begin{abstract}
Real-world image deblurring demands both high-fidelity restoration and computational efficiency, a balance existing methods often struggle to achieve. In this paper, we propose FSM-Net (Frequency-Spatial Multi-branch Network), a highly efficient solution that secured 2nd place in the NTIRE 2026 Challenge on Efficient Real-World Deblurring. FSM-Net pioneers a dual-domain approach: a novel Frequency Attention module explicitly recovers high-frequency structural details via FFT, while a Cross-Gated Vision E-Branchformer at the bottleneck captures global dependencies with linear complexity. To ensure robust convergence, we employ a progressive curriculum training strategy guided by a composite loss function (Multi-Scale Charbonnier, Structural Edge, and Frequency). Evaluated on the RSBlur benchmark, FSM-Net achieves an outstanding 33.144 dB PSNR with only 4.94M parameters and 159.35 GMACs (at 1920x1200 resolution). By effectively pushing the Pareto frontier of efficiency and quality, FSM-Net establishes a strong baseline for resource-constrained image restoration.
\end{abstract}

\section{Introduction}\label{sec:intro}

Real-world image deblurring is a fundamental yet challenging task in computer vision, serving as a crucial preprocessing step for downstream applications like computational photography and autonomous driving \cite{nah2017deep, 10.1007/978-3-030-58595-2_12}. While early attempts heavily relied on perceptual losses \cite{10.1007/978-3-319-46475-6_43} and Generative Adversarial Networks (GANs) \cite{Kupyn_2018_CVPR}, recent advancements have been dominated by Vision Transformers (ViTs) \cite{dosovitskiy2021imageworth16x16words} and large-scale CNNs. Models such as SwinIR \cite{Liang_2021_ICCV}, IPT \cite{IPT_Chen_2021_CVPR}, and MAXIM \cite{Tu_2022_CVPR} have pushed the boundaries of restoration quality. However, their massive parameter counts make them highly impractical for edge deployment.

\begin{figure}[t]
    \centering
    \includegraphics[width=\columnwidth]{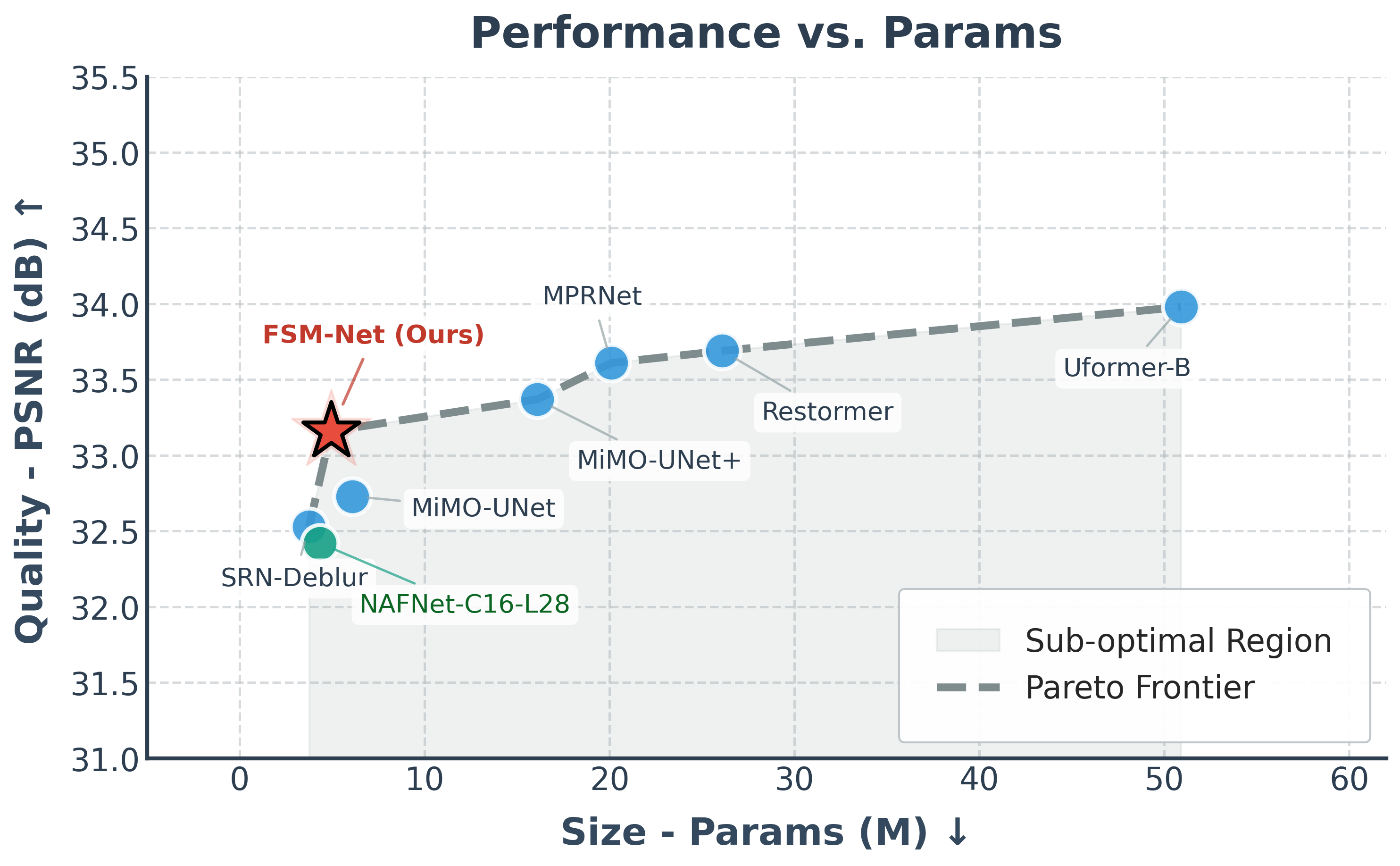}
    \vspace{-4mm}
    \caption{\textbf{Pareto Frontier on the RSBlur Benchmark.} FSM-Net pushes the limits of efficient real-world deblurring, establishing a new optimal trade-off between restoration quality and computational complexity. By explicitly decoupling frequency and spatial features, our 4.94 M parameter model delivers high-fidelity restoration (33.16 dB) that significantly outperforms standard baselines, while remaining vastly more efficient than heavy transformer architectures. FSM-Net secured 2nd place in the NTIRE 2026 Challenge.}
    \label{fig:teaser_pareto}
\end{figure}

The NTIRE 2026 Challenge on Efficient Real-World Deblurring \cite{feijoo2025efficient} highlights a critical paradigm shift: modern networks must deliver high-fidelity restoration while strictly minimizing computational complexity. Existing architectures often struggle with this dilemma, either sacrificing local structural details or relying on heavy self-attention mechanisms that scale quadratically with image resolution \cite{Zamir_2022_CVPR, 10.1007/978-3-031-20071-7_2}. As illustrated by the Pareto frontier in \textbf{Fig. \ref{fig:teaser_pareto}}, balancing global context and local features without severe computational bottlenecks remains a major hurdle.

To tackle this challenge, we propose FSM-Net (Frequency-Spatial Multi-branch Network), an ultra-lightweight architecture built upon NAFNet \cite{10.1007/978-3-031-20071-7_2}. Recognizing that severe motion blur fundamentally acts as a low-pass filter, we introduce a novel FSMBlock equipped with Frequency Attention. By applying Fast Fourier Transform (FFT), our model explicitly suppresses noise and enhances sharp edges directly in the frequency domain. Furthermore, we integrate a Cross-Gated Vision E-Branchformer Block at the network's bottleneck. This module processes information through two parallel branches: a local CNN for precise spatial detail recovery and a 4-head Transposed Attention branch for capturing global context with linear complexity.

Optimized via a robust composite loss, combining Multi-Scale Charbonnier Loss, Structural Edge Loss, and Frequency Consistency Loss, FSM-Net achieves an exceptionally lean profile. At a resolution of $1920 \times 1200$, it requires only 4.94 M parameters and 159.35 GMacs, significantly pushing the Pareto frontier of efficient image restoration.

In summary, our main contributions are as follows:
\begin{itemize}
    \item We introduce FSM-Net, an ultra-efficient architecture that seamlessly integrates spatial and frequency domain processing to optimize the trade-off between restoration quality and computational complexity.
    \item We design the novel FSMBlock with complex-valued Frequency Attention alongside a Cross-Gated Vision E-Branchformer Block, enabling the network to efficiently capture both global receptive fields and sharp local textures.
    \item We demonstrate the superiority of FSM-Net by achieving 33.144 dB PSNR on the NTIRE 2026 public test set, securing 2nd place and establishing a new efficiency benchmark for resource-constrained platforms.
\end{itemize}
\section{Related Work}
\label{sec:relatedwork}

\subsection{Deep Learning for Image Deblurring}
Image deblurring has witnessed a paradigm shift with the rapid evolution of deep neural networks. Early approaches heavily relied on multi-stage or multi-scale Convolutional Neural Networks (CNNs) to progressively restore sharp details. Notable architectures such as SAPHNet \cite{Suin_2020_CVPR}, XYDeblur \cite{Ji_2022_CVPR} MIMO-UNet \cite{Cho_2021_ICCV}, MPRNet \cite{Zamir_2021_CVPR}, and HINet \cite{Chen_2021_CVPR} achieved impressive spatial restoration by leveraging multi-input multi-output designs, cross-stage feature fusion, and instance normalization. Recently, transformers like Stripformer \cite{10.1007/978-3-031-19800-7_9} Vision Transformers (ViTs)\cite{dosovitskiy2021imageworth16x16words} have dominated low-level vision tasks. Restormer \cite{Zamir_2022_CVPR} introduced a highly effective Multi-Dconv Head Transposed Attention (MDTA) to reduce the quadratic complexity of standard self-attention, while Uformer \cite{Wang_2022_CVPR} utilized window-based attention for local feature extraction. Despite their high Peak Signal-to-Noise Ratio (PSNR) on benchmark datasets, these attention-heavy models suffer from massive parameter counts and substantial latency. Recognizing this computational bottleneck, Chen et al. proposed NAFNet \cite{10.1007/978-3-031-20071-7_2}, which replaced non-linear activation functions with simple element-wise multiplications (SimpleGate), achieving competitive performance with significantly reduced FLOPs. Our FSM-Net builds upon the lightweight principles of NAFNet but dramatically enhances its representational capacity through targeted cross-domain feature extraction mechanisms.

\subsection{Frequency-Domain and Hybrid Representations}
Severe motion blur in real-world scenarios inevitably corrupts high-frequency structural details, which are notoriously difficult to recover using purely spatial convolutions. To address this, recent studies have begun exploring the frequency domain for image restoration. Methods utilizing the Fast Fourier Transform (FFT), such as Fast Fourier Convolution (FFC)\cite{NEURIPS2020_2fd5d41e} and Fourier Domain Adaptation (FDA)\cite{Yang_2020_CVPR}, have demonstrated that modulating amplitude and phase can effectively suppress noise and reconstruct sharp edges while enabling robust cross-domain generalization \cite{Mao_Liu_Liu_Li_Shen_Wang_2023, Jiang_2021_ICCV}.\cite{Kong_2023_CVPR}. Furthermore, the integration of local CNNs and global attention mechanisms has proven to be a highly effective hybrid strategy. Architectures such as the Conformer \cite{Peng_2021_ICCV} and E-Branchformer \cite{10022656} in the speech and vision domains demonstrate that processing features through parallel local and global branches yields superior context modeling. Inspired by these advancements, our method introduces the Cross-Gated Vision E-Branchformer block alongside a complex-valued Frequency Attention module, allowing the network to explicitly decouple and learn high-frequency blur degradation patterns without incurring heavy computational overhead.

\subsection{Efficient Restoration and NTIRE Challenges}
The NTIRE workshops have consistently driven the boundaries of image restoration through rigorous challenges \cite{9523065, 9857089, Li_2023_CVPR, Zhang_2024_CVPR, feijoo2025efficient}.
Historically, the pursuit of efficiency has leveraged lightweight convolutions \cite{Sandler_2018_CVPR} and structural re-parameterization\cite{Ding_2021_CVPR}.
Recent editions, including the 2025 Real-World Deblurring challenge\cite{feijoo2025efficient}, demonstrated remarkable inference speeds using highly optimized convolutional backbones. However, these spatial-only architectures often struggle with the long-range, non-uniform blur trajectories inherent in high-resolution imagery. To further push the Pareto frontier in the 2026 challenge, FSM-Net evolves this structural paradigm by integrating a hybrid frequency-spatial design. By decoupling spectral components via Frequency Attention and modeling global context with linear complexity, our model achieves superior restoration fidelity while maintaining the sub-second runtime requirements of mobile ISPs. This architectural advancement allows FSM-Net to handle complex degradations more robustly than prior lightweight spatial-only networks.
\section{Methodology}
\label{sec:method}

To address the strict efficiency constraints of the NTIRE 2026 Challenge, we propose FSM-Net (Frequency-Spatial Multi-branch Network). Building upon the U-shaped NAFNet \cite{10.1007/978-3-031-20071-7_2}, we establish a Pareto balance between restoration quality and computational cost by completely redesigning the feature extraction paradigm to explicitly decouple spatial textures and frequency components. This section details the overall pipeline, core architectural blocks, hybrid loss formulation, and training dynamics.

\begin{figure*}[t]
    \centering
    \includegraphics[width=1\textwidth]{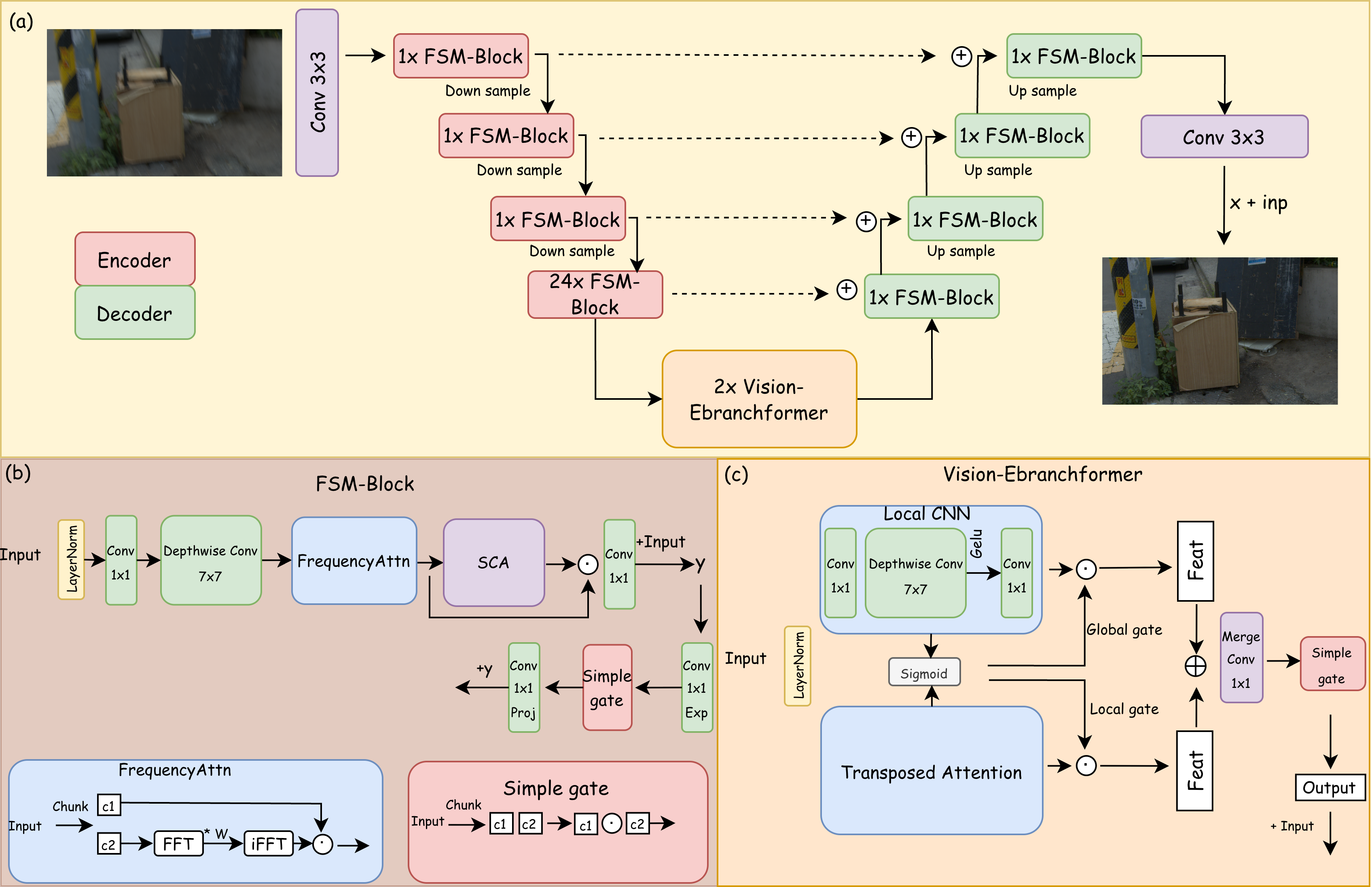}
    \caption{Overall Architecture of FSM-Net. Our model follows an efficient U-shaped hierarchical structure. (a) The global pipeline with 4-level encoder-decoder stages; (b) Detailed schematic of the Frequency-Spatial Multi-branch Block (FSMBlock) featuring Frequency Attention; (c) The Cross-Gated Vision E-Branchformer bottleneck designed for long-range dependency modeling with linear complexity.}
    \label{fig:architecture_full}
\end{figure*}

\subsection{Overall Architecture}
FSM-Net adopts a 4-level encoder-decoder architecture with dense skip connections. Given a blurry image $\mathbf{I}_{blur} \in \mathbb{R}^{3 \times H \times W}$, a $3 \times 3$ convolutional stem projects it into a shallow feature space $\mathbf{X}_0 \in \mathbb{R}^{C \times H \times W}$. 

The encoder progressively downsamples features via strided convolutions (stride 2), doubling channels while halving spatial resolution to extract hierarchical representations. Both the encoder and decoder stages are exclusively populated with our proposed Frequency-Spatial Multi-branch Blocks (FSMBlock) to capture complex degradation patterns. 

At the network bottleneck, where receptive fields are maximized but spatial dimensions are compressed, we introduce a sequence of Vision E-Branchformer Blocks to capture long-range semantic dependencies without inflating MACs. Finally, the decoder progressively upsamples deep features using PixelShuffle\cite{Shi_2016_CVPR} to prevent checkerboard artifacts, culminating in a $3 \times 3$ convolution to predict the clean residual. The final sharp image is obtained via a global skip connection: $\mathbf{I}_{sharp} = \mathbf{I}_{blur} + \mathbf{R}_{pred}$.

\subsection{Frequency-Spatial Multi-branch Block}
Severe motion blur operates not only as a spatial distortion but fundamentally acts as a low-pass filter that severely degrades high-frequency structural details. Standard convolutions are inherently spatially biased and computationally inefficient for modeling such global frequency-domain degradations \cite{NEURIPS2022_91a23b3e, cui2023selective}. Therefore, we design the FSMBlock, which augments the efficient SimpleGate mechanism \cite{10.1007/978-3-031-20071-7_2, 10.5555/3305381.3305478} with a novel Frequency Attention module.

Given an input feature map $\mathbf{X}_{in}$, we first apply Layer Normalization\cite{ba2016layernormalization} followed by a $1 \times 1$ convolution for channel expansion. To capture long blur trajectories efficiently, we apply a depth-wise convolution with a large $7 \times 7$ kernel. Let the resulting intermediate feature be $\mathbf{X} \in \mathbb{R}^{C' \times H \times W}$. We split this feature along the channel dimension into two equal halves: $\mathbf{X}_1$ and $\mathbf{X}_2$. 

To explicitly recover high-frequency information, $\mathbf{X}_2$ is fed into the Frequency Attention module. We first transform $\mathbf{X}_2$ into the frequency domain using the 2D Real Fast Fourier Transform (rFFT):
\begin{equation}
    \mathcal{F}(\mathbf{X}_2)(u, v) = \sum_{h=0}^{H-1} \sum_{w=0}^{W-1} \mathbf{X}_2(h, w) e^{-j 2\pi \left(\frac{uh}{H} + \frac{vw}{W}\right)}
\end{equation}
where $j$ is the imaginary unit. To modulate both the amplitude (contrast) and phase (structure) of the signal simultaneously, we define a learnable complex-valued weight parameter $\mathbf{W}_c \in \mathbb{C}^{1 \times C'/2 \times 1 \times 1}$. The frequency features are element-wise multiplied by this complex weight:
\begin{equation}
    \mathbf{Z}_{fft} = \mathcal{F}(\mathbf{X}_2) \odot \mathbf{W}_c
\end{equation}
The filtered complex signal is then transformed back to the spatial domain via the Inverse Fast Fourier Transform (irFFT), producing the enhanced feature $\mathbf{\tilde{X}}_2 = \mathcal{F}^{-1}(\mathbf{Z}_{fft})$. Unlike real-valued spectral filters that only scale the magnitude, our complex-valued weight $\mathbf{W}_c$ enables simultaneous modulation of both amplitude and phase. Since motion blur fundamentally induces a phase shift and frequency-dependent attenuation, this dual-modulation capability allows FSM-Net to explicitly re-align shifted structural components while amplifying suppressed high-frequency signals. This spectral recalibration is crucial for resolving the ghosting artifacts typically found in non-uniform real-world blur. Finally, the output of the Frequency Attention is fused with the identity spatial branch ($\mathbf{X}_1$) using a non-linear gating mechanism:
\begin{equation}
    \mathbf{X}_{gate} = \mathbf{X}_1 \odot \mathbf{\tilde{X}}_2
\end{equation}

Following this frequency-spatial gating, we introduce a Simplified Channel Attention (SCA) module to recalibrate the channel-wise feature responses and enhance the representational power. We first squeeze the spatial dimensions of $\mathbf{X}_{gate}$ using Global Average Pooling (GAP) to obtain a channel descriptor $\mathbf{s} \in \mathbb{R}^{C'/2 \times 1 \times 1}$:
\begin{equation}
    \mathbf{s}_{c} = \frac{1}{H \times W} \sum_{h=1}^{H} \sum_{w=1}^{W} \mathbf{X}_{gate}(c, h, w)
\end{equation}
This global descriptor is then passed through a $1 \times 1$ convolution to capture cross-channel interactions without any dimensionality reduction. These weights are used to dynamically modulate the gated features: $\mathbf{X}_{sca} = \mathbf{X}_{gate} \odot \mathbf{W}_{sca}(\mathbf{s})$. Finally, $\mathbf{X}_{sca}$ is projected back to the original channel dimension $C$ via a $1 \times 1$ convolution ($\mathbf{W}_{out}$), and added to the residual identity mapping to form the intermediate feature:
\begin{equation}
    \mathbf{X}_{mid} = \mathbf{X}_{in} + \mathbf{W}_{out}(\mathbf{X}_{sca}) \cdot \beta
\end{equation}
where $\beta$ is a learnable layer scaling parameter. 

Subsequently, $\mathbf{X}_{mid}$ is processed by a Feed-Forward Network (FFN) augmented with a SimpleGate to independently model channel-wise transformations. The final block output is obtained via a second residual connection:
\begin{equation}
    \mathbf{X}_{out} = \mathbf{X}_{mid} + \text{FFN}(\mathbf{X}_{mid}) \cdot \gamma
\end{equation}
where $\gamma$ is another learnable scaling factor. By directly modulating the complex wave representations and recalibrating channel dependencies, FSMBlock dynamically suppresses noise and sharpens edges with an $\mathcal{O}(N \log N)$ computational complexity, strictly adhering to the challenge's efficiency requirements.

\subsection{Cross-Gated Vision E-Branchformer Bottleneck}
At the network's bottleneck, capturing global context is paramount for resolving severe, large-scale blur. However, standard Self-Attention mechanisms scale quadratically with spatial resolution $\mathcal{O}(H^2W^2C)$, creating a severe computational bottleneck, while shifted-window variants \cite{Liu_2021_ICCV} often lack sufficient receptive fields. To maintain the ultra-fast inference speed, we propose the Vision E-Branchformer Block, inspired by advancements in speech recognition \cite{10022656}, and adapt it for vision tasks through a novel cross-gating mechanism.

The block splits the normalized input into two parallel, complementary processing branches:
\begin{enumerate}
    \item \textbf{Local CNN Branch:} Employs an inverted bottleneck structure to accurately model fine-grained spatial textures. It consists of a $1 \times 1$ expansion convolution, a $7 \times 7$ depth-wise convolution, a GELU activation\cite{hendrycks2023gaussianerrorlinearunits}, and a final $1 \times 1$ projection convolution, yielding local features $\mathbf{F}_{loc}$.
    
    \item \textbf{Global Attention Branch:} Utilizes a 4-head Transposed Attention module. To inject local inductive bias into the attention mechanism, queries $\mathbf{Q}$, keys $\mathbf{K}$, and values $\mathbf{V} \in \mathbb{R}^{B \times \hat{C} \times HW}$ are first generated using a $3 \times 3$ depth-wise convolution. Instead of computing attention across the massive spatial dimension, we compute the covariance matrix across the channel dimension, inspired by Restormer\cite{Zamir_2022_CVPR}. Before the matrix multiplication, $\mathbf{Q}$ and $\mathbf{K}$ are $\mathcal{L}_2$-normalized along the spatial dimension. The attention is computed as:
    \begin{equation}
        \mathbf{F}_{glob} = \text{Softmax}\left( \hat{\mathbf{Q}} \hat{\mathbf{K}}^T \cdot \tau \right) \mathbf{V}
    \end{equation}
    where $\hat{\mathbf{Q}}$ and $\hat{\mathbf{K}}$ represent the normalized queries and keys, and $\tau$ is a learnable temperature parameter. This transposed formulation reduces the complexity from $\mathcal{O}(H^2W^2C)$ to $\mathcal{O}(HWC^2)$, making global context modeling extremely lightweight.
\end{enumerate}

To synergize the representations from both domains, we introduce a Cross-Gating Mechanism. The features from one branch are utilized to generate attention gates for the other, facilitating mutual information exchange:
\begin{equation}
    \mathbf{F}'_{loc} = \mathbf{F}_{loc} \odot \sigma(\mathbf{F}_{glob})
\end{equation}
\begin{equation}
    \mathbf{F}'_{glob} = \mathbf{F}_{glob} \odot \sigma(\mathbf{F}_{loc})
\end{equation}
where $\sigma(\cdot)$ denotes the Sigmoid activation function. The mutually enhanced features are concatenated, projected via a $1 \times 1$ convolution, and refined through a SimpleGate module. Finally, the output is scaled by a learnable layer scaling parameter before being added to the residual stream.

\subsection{Optimization Strategy}
To optimize FSM-Net for complex real-world degradations, we employ a composite loss function that supervises the network across spatial, structural, and frequency domains. The total objective function $\mathcal{L}_{total}$ is formulated as a weighted linear combination of three distinct loss components:
\begin{equation}
    \mathcal{L}_{total} = w_1 \mathcal{L}_{MSC} + w_2 \mathcal{L}_{Edge} + w_3 \mathcal{L}_{FFT}
\end{equation}
where $w_1, w_2,$ and $w_3$ are hyperparameters that balance the contribution of each domain during the training process.


\paragraph{Multi-Scale Charbonnier Loss ($\mathcal{L}_{MSC}$)}
To capture hierarchical dependencies and enforce spatial fidelity, we utilize a Multi-Scale Charbonnier Loss \cite{413553}. Computed across three scales $s \in \{1, 0.5, 0.25\}$ via average pooling on the predicted ($\mathbf{\hat{I}}$) and ground-truth ($\mathbf{I}$) images, the loss is defined as:
\begin{equation}
    \mathcal{L}_{MSC} = \sum_{s} \frac{\gamma_s}{N_s} \sum_{i=1}^{N_s} \sqrt{(\mathbf{\hat{I}}_{s, i} - \mathbf{I}_{s, i})^2 + \epsilon^2}
\end{equation}
where $\gamma_s$ denotes the scale-specific weight parameter, $N_s$ represents the total number of pixels at spatial scale $s$, $i$ is the spatial pixel index, and $\epsilon = 10^{-12}$ is a small constant ensuring numerical stability. This hierarchical multi-scale supervision effectively encourages the network to restore both coarse global structures and fine-grained local textures.

\paragraph{Structural Edge Loss ($\mathcal{L}_{Edge}$)}
To explicitly enhance the sharpness of recovered boundaries, we introduce an Edge Loss based on the Laplacian operator $\Delta$\cite{10.1098/rspb.1980.0020}. We minimize the Mean Squared Error (MSE) between the extracted high-frequency edges:
\begin{equation}
    \mathcal{L}_{Edge} = \mathbb{E} \left[ \left\| \Delta \mathbf{\hat{I}} - \Delta \mathbf{I} \right\|_2^2 \right]
\end{equation}
By penalizing blurry edges, this component forces the model to prioritize sharp transitions and structural integrity in the restoration process.

\paragraph{Frequency Consistency Loss ($\mathcal{L}_{FFT}$)}
Motivated by the fact that motion blur is fundamentally a frequency-domain degradation\cite{Jiang_2021_ICCV}, we penalize discrepancies in the Fourier space. We apply a Real Fast Fourier Transform (rFFT) to the spatial signals and minimize the $\mathcal{L}_1$ distance between their log-amplitudes:
\begin{equation}
    \mathcal{L}_{FFT} = \mathbb{E} \left[ \left\| \log(1 + |\mathcal{F}(\mathbf{\hat{I}})|) - \log(1 + |\mathcal{F}(\mathbf{I})|) \right\|_1 \right]
\end{equation}
where $\mathcal{F}(\cdot)$ denotes the 2D rFFT operator. The logarithmic scaling ensures that the network remains sensitive to subtle differences in high-frequency spectral components, which are crucial for recovering sharp details in real-world blurry images.

\subsection{Exponential Moving Average}
To ensure training stability and improve generalization without incurring test-time overhead, we integrate Exponential Moving Average (EMA)\cite{morales-brotons2024exponential}. EMA acts as a temporal ensemble, smoothing noisy gradient updates. Let $\theta^{(t)}$ be the model weights at training step $t$, the EMA weights $\theta_{EMA}^{(t)}$ are updated as:

\begin{equation}
    \theta_{EMA}^{(t)} = \alpha \theta_{EMA}^{(t-1)} + (1 - \alpha) \theta^{(t)}
\end{equation}

where the decay rate $\alpha$ is empirically set to 0.999. This mechanism acts as a low-pass filter in the parameter space, dampening weight oscillations and guiding optimization toward wider, \textit{flat minima}. Consequently, EMA significantly enhances the model's robustness against domain shifts, ensuring stable and high-fidelity restoration on unseen test distributions.
\begin{figure*}[t]
    \centering
    \includegraphics[width=1\textwidth]{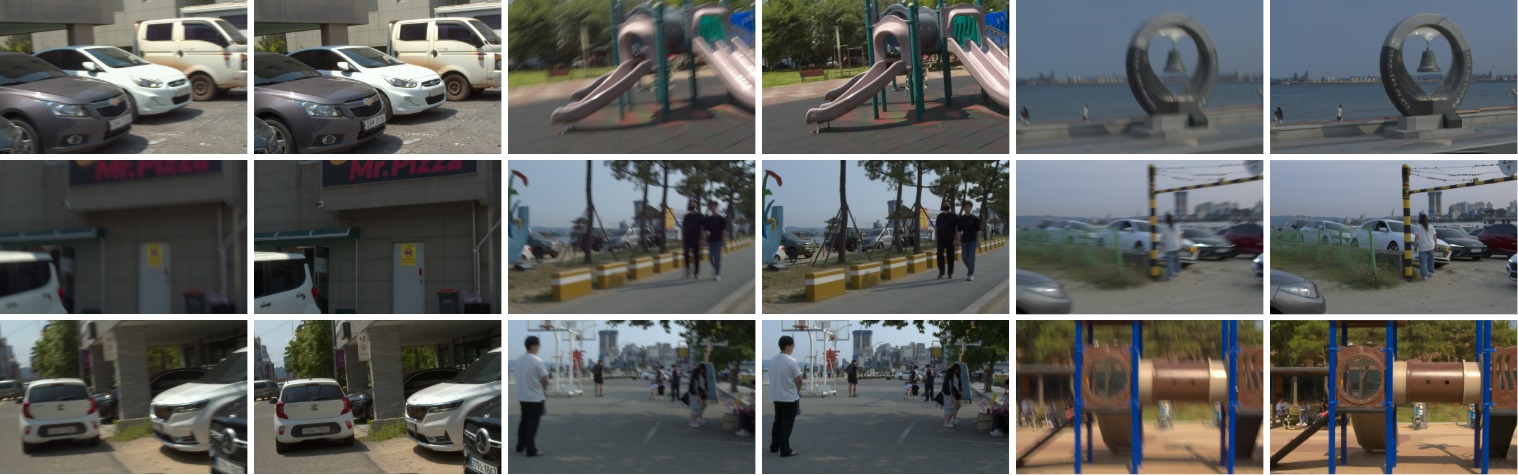}
    \caption{\textbf{Visual results of FSM-Net on the NTIRE 2026 test set.} We present several representative examples to demonstrate the restoration capability of our proposed architecture. FSM-Net effectively recovers sharp structural details and suppresses complex motion artifacts, contributing to our 2nd place ranking on the official public leaderboard. These results highlight the model's efficacy in resolving diverse real-world blurring trajectories. Zoom in for better visualization.}
    \label{fig:visual_demo}
\end{figure*}

\section{Experiments}
\label{sec:exp}

In this section, we provide a comprehensive evaluation of the proposed FSM-Net. We describe the dataset used for the challenge, the intricate multi-stage training details, and present our results in comparison with state-of-the-art methods.

\subsection{Dataset and Evaluation Metrics}
\paragraph{RSBlur Dataset:} We conduct our experiments on the RSBlur dataset \cite{10.1007/978-3-031-20071-7_29}, the official dataset for the NTIRE 2026 Efficient Real-World Deblurring Challenge. To bridge the domain gap between synthetic and real-world degradations, RSBlur employs a sophisticated blur synthesis pipeline. Instead of relying on naive artificial blur kernels, it utilizes real-world sharp image sequences and accurately models the physical camera image formation process, incorporating factors such as sensor noise, saturation, and signal processing pipelines, to synthesize highly realistic blurry/sharp training pairs. The dataset contains diverse indoor and outdoor scenes with complex motion blur patterns. For our training, we utilize the provided training split, consisting of high-resolution image pairs, and evaluate our performance on the full test set.

\paragraph{Evaluation Metrics:} To quantitatively assess the restoration quality, we utilize the Peak Signal-to-Noise Ratio (PSNR), Structural Similarity Index (SSIM) \cite{1284395} and Learned Perceptual Image Patch Similarity (LPIPS)\cite{Zhang_2018_CVPR}. Furthermore, following the challenge mandates, we measure the Runtime per image on a standard GPU to ensure the efficiency of our model. All metrics are calculated in the RGB color space.

\subsection{Implementation Details}
\paragraph{Network Configuration:} The proposed FSM-Net is configured with a base channel width $C=16$. The encoder-decoder depths are set to $[1, 1, 1, 24]$ and $[1, 1, 1, 1]$, respectively, with the number of middle blocks Vision E-Branchformer set to 2 at the bottleneck. This asymmetric design prioritizes deep feature extraction at lower resolutions to maintain a small computational footprint.

\paragraph{Optimizer and Scheduler:} We employ the AdamW optimizer\cite{loshchilov2019decoupledweightdecayregularization} with $\beta_1=0.9, \beta_2=0.999$ and a weight decay of $10^{-4}$. The initial learning rate is set to $10^{-4}$ and is decayed to $10^{-6}$ using a Cosine Annealing scheduler\cite{loshchilov2017sgdrstochasticgradientdescent}. To ensure numerical stability and memory efficiency, we utilize Automatic Mixed Precision (AMP) with \texttt{float16} during the entire training process. Furthermore, an Exponential Moving Average (EMA) with a decay rate of $0.999$ is applied to the model weights to stabilize convergence and enhance generalization.

\paragraph{Multi-stage Curriculum Training:} We implement a 5-phase progressive training strategy over 400 epochs to facilitate convergence from coarse structures to fine textures:
\begin{itemize}
    \item \textbf{Phase 1 (Epochs 0-20):} We utilize a crop size of $512 \times 512$ and optimize solely with the Multi-Scale Charbonnier Loss ($w_1=1.0, w_2=0, w_3=0$) to establish a stable initial spatial mapping.
    \item \textbf{Phase 2 (Epochs 20-120):} Maintaining the $512 \times 512$ resolution, we activate the structural and frequency losses ($w_2=0.01, w_3=0.005$) to begin capturing high-frequency details.
    \item \textbf{Phase 3 (Epochs 120-220):} The crop size is increased to $1024 \times 1024$ to allow the model to learn long-range motion blur trajectories while keeping the loss configuration constant.
    \item \textbf{Phase 4 (Epochs 220-290):} We transition to a $1280 \times 720$ resolution and increase the loss weights to $w_2=0.05$ and $w_3=0.01$ for more aggressive edge and frequency refinement.
    \item \textbf{Phase 5 (Epochs 290-400):} In the final stage, we train on the near-original resolution of $1920 \times 1200$ until convergence, ensuring the model is fully adapted to the high-resolution inference requirements of the challenge.
\end{itemize}
The entire training is conducted on a system equipped with a single NVIDIA RTX 5090 GPU. We utilize a batch size of 4 for Phases 1 through 4. In Phase 5, to accommodate the near-original resolution ($1920 \times 1200$) within the VRAM constraints while maintaining high-fidelity feature maps, the batch size is adjusted to 2.

\subsection{Main Results}
We evaluated FSM-Net on the NTIRE 2026 Efficient Real-World Deblurring Public Test Set, which features complex degradations to test restoration limits under strict computational constraints. As summarized in Table \ref{tab:challenge_results}, FSM-Net secured 2nd place overall with a PSNR of $33.144$ dB and an SSIM of $0.8516$. This validates our core hypothesis: integrating the Frequency Attention module and Cross-Gated Vision E-Branchformer significantly enhances representational capacity without computationally prohibitive network scaling.

Crucially, FSM-Net establishes a highly practical Pareto balance between restoration quality and computational cost. Achieving its peak performance with an optimized latency of just $0.276$s, our model operates significantly faster than other top-ranking submissions reporting runtimes (e.g., Rank 8 at $0.78$s). This sub-second efficiency highlights its strong potential for real-time edge deployment.

Furthermore, FSM-Net outperforms the defending 2025 challenge champion (\textit{licheng}, ranked 4th this year with a PSNR of $32.805$ dB) by a substantial margin of +0.339 dB. This leap underscores the superiority of our dual-domain processing paradigm, demonstrating that our frequency-spatial decoupling effectively breaks the performance ceilings of prior spatial-only architectures.

\begin{table}[t]
\centering
\caption{\textbf{NTIRE 2026 Efficient Real-World Deblurring Challenge Results (Public test Phase).} Methods are ranked by PSNR. The top three identical submissions are grouped as a single entity. Our FSM-Net secures a highly competitive 2nd place overall, offering a highly optimized and fully reported sub-second runtime. The runtime of 0.276s for FSM-Net corresponds to the TTA $\times 4$ ensemble strategy, which was used for the final submission.}
\label{tab:challenge_results}
\setlength{\tabcolsep}{3pt} 
\footnotesize 
\begin{tabular}{@{}clccc@{}} 
\toprule
Rank & Team Name & PSNR$\uparrow$ & SSIM$\uparrow$ & Time (s)$\downarrow$ \\
\midrule
1 & jingjing et al. & 33.390 & 0.8585 & N/A* \\
\textbf{2} & \textbf{RobinLy (Ours)} & \textbf{33.144} & \textbf{0.8516} & \textbf{0.276} \\
3 & weichow & 32.883 & 0.8467 & N/A* \\
4 & licheng & 32.805 & 0.8473 & N/A* \\
5 & zzhlttcyy & 32.657 & 0.8450 & N/A* \\
6 & zhouzhaorun & 32.593 & 0.8422 & N/A* \\
7 & berryzis & 32.322 & 0.8394 & N/A* \\
8 & geometric\_ai & 32.216 & 0.8371 & 0.78 \\
\bottomrule
\end{tabular}
\vspace{-4mm}
\end{table}

\begin{table*}[t]
\centering
\caption{\textbf{Comprehensive Benchmark of State-of-the-Art Methods Across Multiple Datasets.} For cross-dataset evaluation (RealBlur-R, RealBlur-J, and GoPro), FSM-Net was fine-tuned for only 5 epochs from the RSBlur pre-trained weights. ``-'' indicates that the metric or result is not explicitly reported or available for that specific configuration. FSM-Net achieves an exceptional balance between high-fidelity restoration and computational efficiency across diverse real-world and synthetic degradation profiles.}
\label{tab:comprehensive_benchmark_large}
\renewcommand{\arraystretch}{1.1} 
\begin{tabular}{lccccc}
\toprule
\multirow{2}{*}{Method} & \multirow{2}{*}{Param (M)$\downarrow$} & \multicolumn{4}{c}{Dataset Performance (PSNR $\uparrow$ / SSIM $\uparrow$)} \\
\cmidrule(lr){3-6}
 &  & RSBlur & RealBlur-R & RealBlur-J & GoPro \\
\midrule
SRN-Deblur \cite{8578951} & \textbf{3.76} & 32.53 / 0.8398 & \textbf{38.65} / \textbf{0.9652} & \textbf{31.38} / \textbf{0.9091} & 28.36 / 0.9150 \\
NAFNet-C16-L28\cite{10.1007/978-3-031-20071-7_2} & 4.35 & 32.42 / 0.8400 & - & - & - \\
DeblurGAN-v2 \cite{Kupyn_2019_ICCV} & 6.09 & - & 36.44 / 0.9347 & 29.69 / 0.8703 & - \\
MiMO-UNet \cite{Cho_2021_ICCV} & 6.10 & 32.73 / 0.8457 & - & - & - \\
MiMO-UNet+ \cite{Cho_2021_ICCV} & 16.1 & 33.37 / 0.8560 & - & - & - \\
MPRNet \cite{Zamir_2021_CVPR} & 20.1 & 33.61 / 0.8614 & - & - & 30.96 / 0.9390 \\
Nah \etal \cite{nah2017deep} & 22.6 & - & 32.51 / 0.8406 & 27.87 / 0.8274 & 29.08 / 0.9135 \\
Restormer \cite{Zamir_2022_CVPR} & 26.1 & 33.69 / 0.8628 & - & - & \textbf{31.22} / \textbf{0.9420} \\
Uformer-B \cite{Wang_2022_CVPR} & 50.9 & \textbf{33.98} / \textbf{0.8660} & - & - & 30.83 / 0.9520 \\
\midrule
FSM-Net (Ours) & 4.94 & 33.16 / 0.8533 & 36.95 / 0.9585 & 29.45 / 0.8840 & 30.60 / 0.9068 \\
\bottomrule
\end{tabular}
\end{table*}

\paragraph{Comprehensive SOTA Benchmark and Cross-Dataset Generalization}
To rigorously evaluate FSM-Net, we present a comprehensive benchmark against state-of-the-art (SOTA) deblurring methods. This evaluation is twofold: assessing restoration efficacy on the official RSBlur public full test set (3,360 real-world image pairs) and analyzing cross-domain generalization on three diverse external datasets (RealBlur-R, RealBlur-J, and GoPro). 

As detailed in Table \ref{tab:comprehensive_benchmark_large}, FSM-Net achieves high-fidelity restoration on the RSBlur benchmark with a PSNR of 33.16 dB and an SSIM of 0.8533. It is crucial to emphasize that these metrics are obtained using a direct, single forward pass without the aid of any Test-Time Augmentation (TTA). Notably, FSM-Net delivers this competitive, unaugmented performance while requiring a mere fraction of the computational footprint (4.94M parameters) compared to heavy SOTA architectures like MPRNet\cite{Zamir_2021_CVPR} or Uformer-B\cite{Wang_2022_CVPR}.

Furthermore, to validate cross-domain adaptability, we initialized FSM-Net with our best RSBlur pre-trained weights and performed a lightweight fine-tuning of merely 5 epochs on the external benchmarks. Despite this minimal adaptation phase, FSM-Net exhibits rapid convergence and robust generalizability. On the real-world RealBlur-R and RealBlur-J\cite{10.1007/978-3-030-58595-2_12} datasets, our method achieves 36.95 dB and 29.45 dB respectively, remaining highly competitive with baselines specifically trained on those domains. Simultaneously, its stable performance on the synthetic GoPro dataset underscores that the representational capacity established by our frequency-spatial decoupling paradigm transfers effectively to unseen motion blur trajectories without severe overfitting.

\paragraph{Visual Performance Analysis}
Figure \ref{fig:visual_demo} presents a qualitative evaluation on the NTIRE 2026 test set, demonstrating FSM-Net's capability to mitigate severe motion artifacts. Specifically, our Frequency Attention module accurately reconstructs dense high-frequency details, such as legible text on the "Mr.Pizza" signage and sharp license plates, by explicitly eliminating ghosting artifacts. For large-scale geometries subjected to long blur trajectories (e.g., playground equipment and the waterfront monument), the Cross-Gated Vision E-Branchformer successfully restores structural integrity without warping. Furthermore, our Multi-Scale Charbonnier Loss ensures background regions like the sky and foliage remain perceptually smooth, preventing the over-sharpening halos common in traditional methods. These qualitative improvements directly corroborate our quantitative success (2nd place, PSNR: 33.144 dB, SSIM: 0.8516).

\begin{table}[htbp]
\centering
\caption{\textbf{Architectural Ablation.} Evaluated on 3,360 images after a 20 epoch high resolution training schedule.}
\label{tab:arch_ablation}
\setlength{\tabcolsep}{1.5pt} 
\scriptsize 
\begin{tabular}{@{}lccccccc@{}} 
\toprule
Model & FAttn & Ebranch & Param (M) & MACs (G) & PSNR$\uparrow$ & SSIM$\uparrow$ & LPIPS$\downarrow$ \\
\midrule
Baseline & \texttimes & \texttimes &\textbf{4.35} & \textbf{146.33} & 31.39 & 0.8175 & 0.3557 \\
+ FAttn & \checkmark & \texttimes & 4.66 & 155.66 & 31.57 & 0.8217 & 0.3494 \\
+ Ebranch & \texttimes & \checkmark & 4.93 & 159.35 & 31.60 & 0.8216 & 0.3478 \\
\textbf{FSM-Net} & \checkmark & \checkmark & 4.94 & 159.35 & \textbf{31.83} & \textbf{0.8282} & \textbf{0.3456} \\
\bottomrule
\end{tabular}
\end{table}

\subsection{Ablation Study}

\paragraph{Architectural Components Ablation:}
To rigorously evaluate the individual contributions of our proposed modules under strict computational constraints, we conducted an accelerated 20-epoch ablation study at high resolution ($1920 \times 1200$) using the 3,360 sample full test set. As shown in Table \ref{tab:arch_ablation}, the spatial-only baseline (NAFNet)\cite{10.1007/978-3-031-20071-7_2} establishes a foundation at 31.39 dB PSNR. Integrating the Frequency Attention (FAttn) module yields a noticeable improvement to 31.57 dB, demonstrating the necessity of explicit spectral recalibration for mitigating high-frequency motion artifacts. Alternatively, deploying the Vision E-Branchformer (Ebranch) enhances spatial long-range dependency modeling, elevating the performance to 31.60 dB. 

Crucially, when both pathways are synergized in the full FSM-Net, the restoration quality peaks at 31.83 dB with a highly competitive LPIPS of 0.3456. Notably, this combined improvement (+0.44 dB over the baseline) strictly exceeds the sum of their individual marginal gains (+0.18 dB and +0.21 dB). This non-linear performance boost empirically proves that our frequency and spatial modules are highly complementary, successfully capturing orthogonal degradation features rather than redundant information. From an efficiency standpoint, this substantial perceptual and quantitative leap incurs only a marginal computational overhead (+0.59 M parameters, +13.02 GMacs). This firmly validates that our dual-domain design fundamentally expands the network's representational capacity without violating the strict boundaries of efficient deployment.
\begin{table}[t]
\centering
\caption{\textbf{Ablation Study on Test-Time Augmentation (TTA) Strategies.} The trade-off between restoration quality (PSNR/SSIM) and computational efficiency (Latency) evaluated on the RSBlur full test set. Latency is measured on a single NVIDIA RTX 5090 GPU.}
\label{tab:tta_ablation}
\setlength{\tabcolsep}{2.5pt} 
\footnotesize 
\begin{tabular}{llccc} 
\toprule
TTA & Transforms & PSNR$\uparrow$ & SSIM$\uparrow$ & Time (s)$\downarrow$ \\
\midrule
$\times 1$ & None & 33.006 & 0.8500 & \textbf{0.069} \\
$\times 2$ & + H-Flip & 33.105 & 0.8510 & 0.138 \\
$\times 3$ & + V-Flip & 33.135 & 0.8515 & 0.207 \\
$\times 4$ & + D-Flip & \textbf{33.144} & \textbf{0.8516} & 0.276 \\
\bottomrule
\end{tabular}
\end{table}

\paragraph{Test-Time Augmentation (TTA) Trade-offs:}
To maximize restoration quality within the sub-second constraint, we evaluate four progressive TTA configurations (Table \ref{tab:tta_ablation}). The baseline (TTA $\times 1$) achieves an ultra-fast 0.069s latency with a competitive 33.006 dB PSNR. Progressively incorporating horizontal (TTA $\times 2$) and vertical (TTA $\times 3$) flips yields consistent PSNR gains. For the full ensemble (TTA $\times 4$, including a diagonal flip), we implement a weighted blending strategy (0.4 for the original, 0.2 for each augmentation) rather than a naive mean. This preserves the original spatial cues, achieving a peak PSNR of 33.144 dB at 0.276s. Ultimately, this demonstrates that FSM-Net can dynamically adapt to strict latency budgets (0.07s–0.28s) without any architectural modifications.

\section{Conclusion}
\label{sec:conclusion}

In this paper, we presented FSM-Net, an ultra-efficient frequency-spatial multi-branch architecture tailored for real-world image deblurring. By synergizing spatial-domain convolutions with complex-valued Frequency Attention, our model establishes a new Pareto frontier, achieving a superior balance between high-fidelity restoration and low computational overhead. With only 4.94 M parameters and a lean complexity of 159.35 GMacs at $1920 \times 1200$ resolution, FSM-Net addresses the critical requirements for high-resolution processing in resource-constrained environments. The proposed method demonstrated outstanding performance during the NTIRE 2026 challenge, currently securing 2nd place on the public leaderboard with a PSNR of 33.144 dB. FSM-Net thus establishes a strong, efficient baseline for future research in resource-constrained high-fidelity image restoration.
\clearpage
{
    \small
    \bibliographystyle{ieeenat_fullname}
    \bibliography{main}
}


\end{document}